# Hard Math – Easy UVM

Pragmatic solutions for verifying hardware algorithms using UVM

Mark Litterick, Verilab, Munich, Germany (mark.litterick@verilab.com)

Aleksandar Ivankovic, Verilab, Belgrade, Serbia (aleksandar.ivankovic@verilab.com)

Bojan Arsov, Verilab, Belgrade, Serbia (bojan.arsov@verilab.com)

Aman Kumar, Infineon Technologies, Dresden, Germany (aman.kumar@infineon.com)

*Abstract*— **This paper presents pragmatic solutions for verifying complex mathematical algorithms implemented in hardware in an efficient and effective manner. Maximizing leverage of a known-answer-test strategy, based on predefined data scenarios combined with design-for-verification modes, we demonstrate how to find and isolate concept and design bugs early in the flow. The solutions presented are based on real project experience with single chip radar sensors for a variety of applications. The verification environments supporting the presented strategies are based on SystemVerilog and the Universal Verification Methodology.**

## I. Introduction

This paper presents pragmatic solutions for verifying complex mathematical algorithms implemented in hardware in an efficient and effective manner. The solutions presented are based on real project experience with single chip radar sensors for a variety of applications [1]. The verification environments supporting the presented strategies are based on SystemVerilog and the Universal Verification Methodology (UVM).

One of the recognized problems with the verification and debug of such complex digital signal processing is that defects are hard to characterize and localize. Building a fully functional mathematical model is cumbersome and time consuming and tends to not isolate the problem well enough or early enough, even when the answer is clearly incorrect.

We demonstrate a novel approach which leverages a known-answer-test (KAT) strategy based on a limited set of predefined data scenarios with predicted results. This is combined with a design-for-verification (DFV) concept utilizing built-in hardware test-modes to facilitate incremental algorithm development in isolation from full chain digital signal processing (DSP). The KAT techniques are extended to validate the full algorithm chain using constrained random stimulus and embedded modeling, which enables us to explore the effect of the configuration randomization on results while reusing the existing data scenarios. These techniques enable verification of hardware-based discrete fast Fourier transform (FFT), constant false alarm rate (CFAR), moving target indication (MTI) functions which in turn support presence, range, motion, and angle detection algorithms.

## II. Architecture

The techniques presented in this paper are not restricted to a specific application, however they are presented in the context of mathematical algorithms for digital signal processing embedded in a single-chip radar device. A simplified block diagram of such a device under test (DUT) and the UVM testbench environment is shown in Figure 1. A detailed discussion of the entire device and testbench environment is not within the scope of this paper, however more details are given in [5]. Here we will focus on the DSP part of the device, whose primary function is to provide autonomous motion and target detection while the host environment is in low-power mode or otherwise occupied. On successful detection of enough consecutive targets meeting the configured criteria, the device can interrupt the host and it can perform additional signal processing to decipher gestures or other details from dedicated radar operations making use of the main data-path via the FIFO.

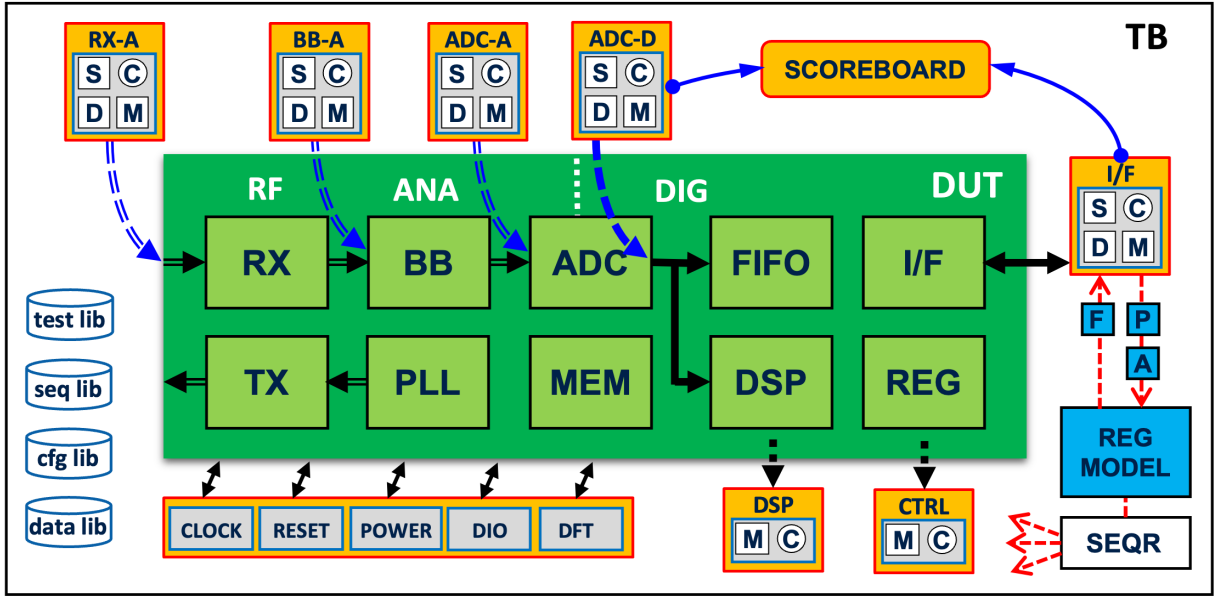

Figure 1 – Block Diagram

A more detailed diagram of the hardware DSP part of the chip is shown in Figure 2. Here we can see each algorithm step in the DSP chain and the interaction between the blocks as well as shared resources such as memory and registers.

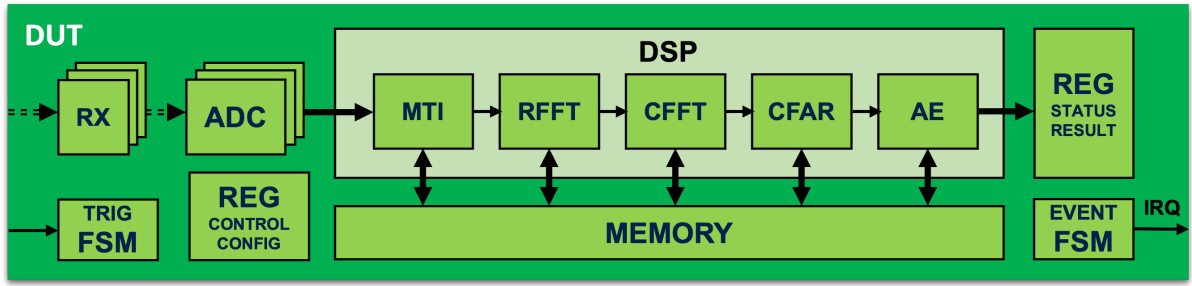

Figure 2 – Hardware DSP Chain

The DSP algorithm steps are applied consecutively to receiver (RX) data streams coming from an array of analogue to digital convertors (ADCs), each of which digitizes the output from orthogonal RF antennas after baseband analog processing. Each RX channel is processed independently and then compared to determine range and movement, and then compared with one another to extract angle and direction information. The data are logically organized into bursts of samples and are stored in fixed-point format. The main algorithm steps are performed in the following order. Note that the descriptions are for guidance only, it is not appropriate to provide proprietary details of the algorithms and it is not necessary to understand the details to appreciate the concepts presented in this paper.

A. Moving Target Indication (MTI)

MTI is a radar technique to identify a moving target against background clutter. It calculates the arithmetic difference between consecutive bursts.

B. Real Fast Fourier Transform (RFFT)

RFFT is used to identify the range of potential targets. It is performed on all bursts produced by the MTI step, resulting in M-1 RFFTs with N/2 bins.

*C. Complex Fast Fourier Transform (CFFT)*

CFFT is used to identify the velocity of potential targets. It is performed on all samples of the RFFT, across bursts, resulting in N/2 CFFTs with M-1 bins each. The data is arranged into a matrix of complex numbers with dimension N2×(M-1).

*D. Constant False Alarm Rate (CFAR)*

CFAR is an adaptive algorithm used in radar systems to detect target returns against a background of noise, clutter, and interference. CFAR is performed on each CFFT bin. It is a complex algorithm that uses CORDIC functions to calculate the magnitude of the range bins and compares the results against target zones in an iterative manner. The algorithm provides hooks for filtering and guard-banding (to minimize false target detection). Multiple targets can be detected and prioritized.

*E. Angle Estimation (AE)*

In a typical device there are at least three RX antennas with different physical orientation within the package, hence it is possible to compute the azimuth and elevation angles from the corresponding channel data using an angle estimation algorithm. The CFFT representations of the potential targets are processed using matrix arithmetic and CORDIC operations to identify the angle and direction of target movement.

### III. DESIGN-FOR-VERIFICATION

Design-for-verification is a technique for improving the effectiveness of a verification environment by implementing appropriate test structures into the actual design that can be used to provide better control or observability of the overall operation. For DFV to be most effective, these additional test structures should be non-obtrusive and do as little interference as possible. In effect we do not want to introduce the scope for more bugs to be inserted by conceiving convoluted test structures. Typically, these test structures can also be used by post-silicon test-engineering or validation teams during the characterization and debug phase as well, but that is not their primary intent.

One effective DFV mode for the DSP chain is to provide lock-step control over the execution of each complete algorithm step (not each clock cycle). We worked with our design colleagues to ensure that we could trigger each step individually and in isolation from the other steps. In addition, we required the ability to preload memory state prior to each step and read out the results from memory after each step. This means that the DSP algorithm state machine had to start and stop individual steps on-demand during the test mode, and not automatically continue processing and overwriting memory as is does in normal operation. Such an approach allows us to comprehensively test each step, in isolation from the others, allowing much better bug detection and isolation as well as supporting linear design development. For example, even if there are bugs in earlier steps, we can verify subsequent steps in the chain. The mechanism also allows for several steps to be combined in series and, of course, full chain operation when required. The supported algorithm steps and associated data scenario files (discussed in the following section) are show in Figure 3

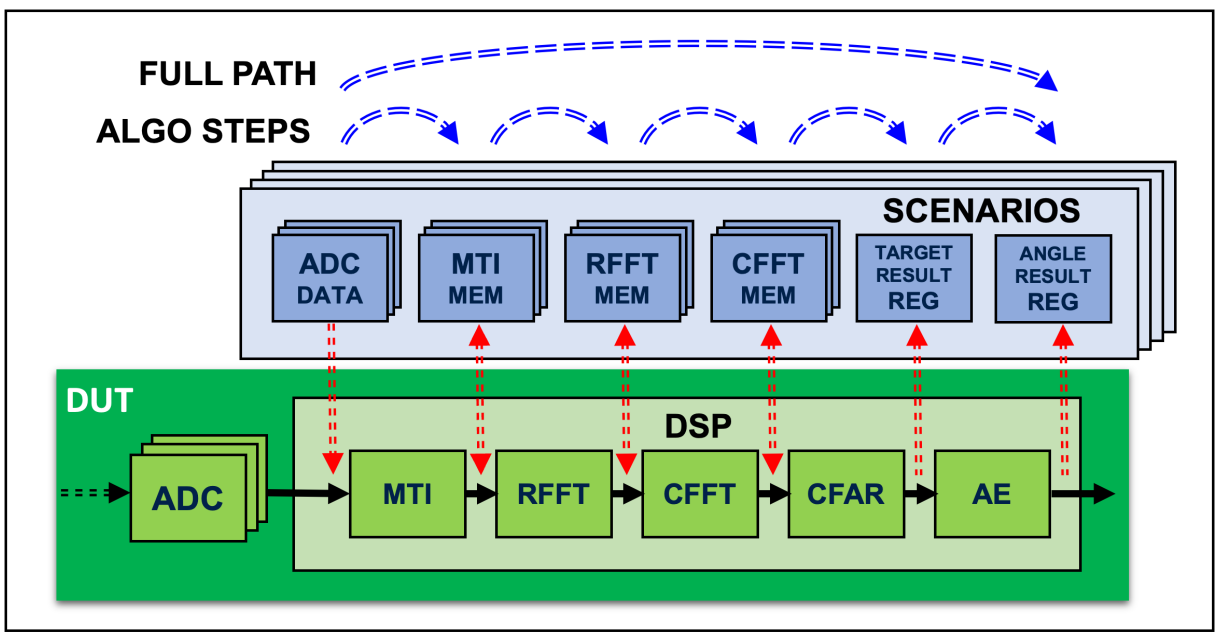

Figure 3 – DFV Steps and Data Scenarios

## IV. Data Scenarios

In the context of the radar chip, a data scenario is essentially a set of receive data values derived from reflections of the transmitted signal. After analog processing the reflected data contains information about potential targets as well as noise and clutter. For verification purposes the data set also includes expected results for a given configuration and, to support the DFV techniques discussed in the previous section, we also have intermediate data results for each step in the algorithm chain.

Typically, the radar data at the ADC outputs are not constrained random values, instead it is synthesized in a MatLab environment or captured from real-world observations. The results for each algorithm step are calculated by an external mathematical analysis tool such as MatLab and encapsulated in separate files. The results after the CFAR and AE steps are contained in registers and are also predicted by the MatLab flow. Each data scenario from Figure 3 consists of a set of files containing the following information:

- Scenario description and intent, configuration settings (REG), expected results (CFAR, AE)
- Data file with raw ADC data radar samples (for all available RX channels)
- Memory files with data values and locations for each step (MTI, RFFT, CFFT)

Note that to make use of the intermediate result data we need to know implementation details about the data structures in memory (values, format, location, etc.), even though this is not a primary verification requirement for the device. This slight disadvantage is, in our opinion, more than compensated for by the advantages that the DFV techniques bring to isolation and debug.

One of the key aspects of the techniques proposed in this paper is that we do not need to burden the project with an exhaustive set of data scenarios. Rather, we work with the concept and application colleagues to get a minimal set of data scenarios that can be used to verify accuracy and performance of the RTL implementation for each algorithm step and the entire DSP chain under different configurations. In the current project we started with less than ten data scenarios and added a couple more during the project development. Some examples of data scenarios are described in abstract terms below (for guidance only):

- Strong single target (in low-noise environment)
- Multiple targets with different characteristics (range, velocity, direction)

- High-noise environment (where target detection depends on filter and declutter settings)

## V. TESTCASE STYLES

We define a variety of test types which all use the data scenarios described in the previous section. These tests include test-modes for the incremental algorithm steps, limited but carefully chosen randomization of sequence-based tests for the full chain, and support for externally defined system use-cases running on the UVM environment. These test features are isolated and combined using the techniques described in [2] and [3]. The test types fall into the following categories:

### A. DFV Step KAT

For each algorithm step there is a dedicated KAT which runs with all available data scenarios. Expected results are taken directly from data scenario. The source data and expected results depends upon the algorithm step (e.g. MTI: ADC to MEM, FFT: MEM to MEM, CFAR: MEM to REG). These tests were executed first to debug each algorithm step independently. The focus here was to identify fundamental issues with each individual algorithm step before combining or debugging in a full path chain.

### B. Full Path KAT

For each available data scenario there is dedicated KAT test for the full chain operation. Expected results are taken directly from data scenario. The full path KAT tests were implemented and debugged as the second stage of verification. In this case all algorithm steps had to work together, and we were focused on overall DSP results as well as integration issues such as resource conflicts, data formatting and arbitration.

### C. DSP Feature

For each high-level application DSP feature (e.g. motion detect, target acquisition, angle estimation), there is a constrained-random test using source data from each appropriate scenario but varying all relevant configuration parameters. In this case the results are calculated by models (see Section VI), since the configuration space goes beyond the predicted results in the data scenarios (i.e. with the same source data but different configuration setup we get different results). We were able to postpone full implementation of these tests and corresponding modeling until much later in the project, but initial versions were possible after the Full Path KAT tests were working.

### D. Application Usecase

The testcases from the verification team are supplemented by a small set of directed tests from other stakeholders (e.g. concept and application engineering). These independent tests also run on the UVM environment but do not typically use our configuration sequences or helper methods. Source data is taken from an appropriate data scenario, but expected results are calculated externally and hard coded into the test. These tests were supported throughout the project, but only became effective after the Full Path KAT tests were successfully passing.

## VI. ROLE OF MATHEMATICAL MODELING

Once we have established the fundamental correctness of the implemented algorithms with all major modes of operation and identified and fixed all the primary bugs, the testbench is then able to evolve to the next stage containing a fully functional mathematical model to predict results. The goal is to explore the algorithm space more thoroughly with verification closure on all aspects of data and configuration stimulus, checking and coverage.

At this point it should be stressed that despite the lack of a local model for algorithm implementation, we were able to detect many design and specification defects within the evolving DUT. This would have been very difficult to achieve within the same timeframe if we had opted for the implementation specific modelling approach from the start. Of course, we still required the high-level mathematical modelling from the MatLab team to generate the scenarios, but that model is not corrupted by implementation specific details or cycle-accurate limitations. Furthermore, debugging model behaviour was relatively straightforward in an environment with working KAT tests and RTL that was proven for this subset of functionality.

Initial floating-point models were implemented in SystemVerilog and were derived directly from the high-level specification. This enabled the verification team to provide independent interpretation of the specification and remain reasonably decoupled from the algorithm development team. However, this independence comes at a price in terms of model coding and debugging. Later in the project, when a fixed-point model became available from the algorithm team, we switched to using DPI calls to C methods which were exported directly from MatLab. Although this removed a degree of independence, the MatLab algorithms had evolved to the stage where they were effectively the golden reference model and had superseded the higher-level description in the specification.

## VII. Error Tolerance & Uncertainty

It is inevitable that with differing implementation details, the model and the hardware will accumulate calculation errors (rounding, precision, etc.) in different ways (implementation of arithmetic in software, arithmetic modules in the hardware such as multipliers, dividers, the CORDIC, etc.). This potentially introduces uncertainty in our checking, which needs to be addressed.

For both the models and the hardware implementation, it is possible to quantify the calculation errors, and in the case of the hardware, they should be known beforehand by the design and concept teams. This quantification of errors enables the establishment of tolerance in the testbench checks, including both intermediate algorithm step checking and final target magnitude checking. The uncertainty was more significant with earlier versions of the floating-point models, but reduced significantly as we evolved towards fixed-point models which more closely matched the RTL implementation.

However, some testbench uncertainty challenges cannot be solved just by adding tolerance. For example, if two magnitudes computed by the model (due to imperfect precision) are close enough together to be modelled by almost the same value and one of them is determined to be the strongest target, it can happen that the hardware implementation gives a different answer, where it would be extremely difficult to determine which answer is correct (where both are acceptable). We attempt to avoid these conflicts by constraining data scenario content, and where appropriate configuration parameters, such that the stimulated target magnitudes differ enough to avoid uncertainty in results.

## VIII. UVM Implementation

This section provides an overview of the UVM testbench implementation details associated with the techniques discussed in this paper. Unfortunately, a full discussion on all UVM testbench details is outside the scope of this paper. Nonetheless, we hope that the implementation details provided here can be interpreted by readers with existing UVM knowledge in order to understand the low-level details. Others should be able to treat the content as pseudo-code or skip the section entirely.

### A. Scenario Selection

Many of the tests are constrained to randomly select an appropriate data scenario. To provide additional control at regression level, we implemented a mechanism to constrain the data scenario from the command line using the standard SystemVerilog *$plusargs* system method as shown below (1).

```
class seq_base_kat extends tb_base_seq;

    string     scenario_s; // selected DSP scenario
  rand scenario_t m_scenario; // derived DSP scenario

  void'($value$plusargs("SCENARIO=%s", scenario_s); // in new()

  constraint scenario_c { // random for empty or unrecognized string
    (scenario_s == "single") -> m_scenario == SCENARIO_SINGLE;
    (scenario_s == "multi")  -> m_scenario == SCENARIO_MULTI;
    (scenario_s == "noise")  -> m_scenario == SCENARIO_NOISE;
  }
```
(1)

## B. File Reading

The format used for the data scenario file set content (2) is chosen so that we can perform easy extraction of the values using the standard SystemVerilog `$readmemh` system function as shown below (3). Note that several data formats are required for the different data representations used in the DSP chain, so we provide a set of helper methods in the base classes to encapsulate this (e.g. `get_rx_data()`, `get_adc_data()`, etc.).

```
// data scenario files : whitespace, single and multiline comments supported
@123   E3B0 // example for Q12,4 format
@EC50 FF8174 // example for Q0.23 format
```
(2)

```
typedef bit[15:0] adc_data_t [int unsigned]; // 16-bit assoc array, index by address

function adc_data_t get_adc_data(string filename);
  adc_data_t adc_data;
  $readmemh(filename,adc_data); // read all data into array in single step
  return adc_data;
```
(3)

## C. ADC Streaming

The ADC values for the DSP source in the corresponding data scenarios are driven onto the internal ADC result signals using the data streaming techniques described in [4]. The testbench does not interfere with the state-machine or control operation, but rather overloads the result values using a force, whenever a result strobe is indicated. This data streaming operation is under control of the sequence hierarchy (4), with the driver providing the forcing mechanism (5) via the interface as shown below (6). Note that if the sequence is not called, the interface signal does not change, and the internal signals are not forced to any value or otherwise disturbed.

```
class madc_digital_file_seq extends tb_base_seq; // in seq_lib
  adc_digital_seq_item adc_item;
  virtual task body();
    adc_data_t adc_data = cfg.get_adc_data(filename);
    `uvm_do_on_with(adc_item, p_sequencer.madc_digital_sequencer[0], { // do item
      foreach (adc_item.m_data[i]) adc_item.m_data[i] == adc_data[i];
```
(4)

```
task adc_digital_driver::drive(...);
  foreach (req.m_data[i]) begin
    @(posedge vif.result_strobe);
    vif.drive_result <= req.m_data[i]; // drive interface signal
```
(5)

```
module tb_top;
  always @(adc_digital_if.drive_result)
    force `ADC_RESULT_PATH = adc_digital_if.drive_result; // force internal signal
```
(6)

## D. Memory I/O

UVM sequences are used to provide all memory input/output stimulus and result checking. These sequences make use of the helper methods described in VIII.B, but also apply another set of helper methods to determine the memory mapping for each step in the chain. Each sequence allows the option to execute via the UVM frontdoor (resulting in physical signal protocol on the serial interface shown in Figure 1) or backdoor (which deposits or extracts data directly from the memory cells in the DUT). Backdoor access is typically used for the DFV Step KAT tests described in V.A and frontdoor access is used for all other testcases.

## E. Register I/O

UVM sequences are used to provide all configuration stimulus and status register check operations via an automatically generated UVM register model which is integrated into the environment using frontdoor, predictor and adaptor components. With the exception of the application usecases, the test sequences do not contain direct references to register fields, but make use of a comprehensive sequence library which follows the principles described in [2] and [3]. The sequence library is composed of configuration, control, memory, status/result and action sequences (e.g. trigger radar operation, wait for interrupt).

*F. Results Checking*

UVM sequences are used to encapsulate all necessary status register and memory result checks at the test sequence level in the testbench. For the KAT results, the expected memory and register results are extracted from the corresponding data scenario file using the helper methods described in VIII.B. For the random feature tests, the expected results are calculated using the models described in VIII.G. In addition, there is a family of helper methods for calculating the performance (duration) of the algorithm steps (and full chain), and signal protocol checks embedded in the interface of a passive verification component which supervises the control signals for the DSP operation.

*G. Model Implementation*

The testbench modeling for the DSP algorithms is implemented in a family of methods which are encapsulated inside the DSP UVC. These methods use configuration settings and selected scenario data to calculate the expected results in both the memory and in the result and status registers. The methods are stand-alone but can be cascaded to combine multiple steps in the DSP chain. Some examples of the method declarations are shown below (7):

```
class ifx_dsp_config extends uvm_object;
  // calculate MTI result in MEM
  function void calc_mti_result(ref sample_array_t data);
  // calculate RFFT result in MEM
  function void calc_range_fft_result(ref sample_array_t data);
  // calculate DFFT result in MEM
  function void calc_doppler_fft_result(ref sample_array_t data);
  // calculate CFAR result in REG
  function target_list_t calc_cfar_result(input sample_array_t data);
  // calculate AE result in REG
  function angle_list_t calc_angle_result(input sample_array_t data);
```
(7)

The pure mathematical calculations are isolated from type conversion and memory distribution methods to ensure a compact and reusable code structure. Models implemented natively in SystemVerilog make extensive use of the generic math system functions (e.g. $\$sin$, $\$cos$, $\$atan2$, etc.) for the calculations. Alternative models make direct use of MatLab C functions by calling the high-level methods via the DPI (8) (9).

```
package dsp_pkg;
  import "DPI-C" function chandle DPI_cfft_initialize(… chandle …);
  import "DPI-C" function void    DPI_cfft_calculate(… pnts … r_i, i_i … r_o, i_o);
  import "DPI-C" function void    DPI_cfft_terminate(… chandle …);
```
(8)

```
chandle objhandle = DPI_cfft_initialize(objhandle);
DPI_cfft_calc(objhandle, num_points, real_i, imag_i, real_o, imag_o);
DPI_cfft_terminate(objhandle);
```
(9)

*H. Test Style*

With the sequence library and helper methods described in the previous sub-sections, we can construct compact test descriptions in the sequences (11) (13) for all the test types described in Section V using base sequence layers (10) (12) as shown below.

```
class seq_dsp_base_kat extends tb_base_seq;
  rand scenario_t  m_scenario;   // scenario selection
  rand algorithm_t m_algorithm;  // algorithm selection
  task load_xxx_data(scenario_t scenario);    // adc, mti, rfft, dfft
  task check_xxx_results(scenario_t scenario); // mti, rfft, dfft, cfar, ae
  task body();
    `uvm_do(init_seq)                   // init clock, power and reset
    `uvm_do_with(config_dsp_seq,{...})  // execute dsp config seq
    load_xxx_data(scenario);            // load input data from scenario file
    `uvm_do_with(dsp_step_seq,{...})    // trigger dsp step with calculated timeout
```
(10)

```
        check_xxx_results(scenario);        // check output data from scenario file
```
 

```
class seq_basic_xxx_kat extends seq_usecase_dsp_base; // where xxx = algorithm step
  constraint xxx_c {m_algorithm == ALGO_XXX;} // selected DSP step on all scenarios
```
(11)

```
class seq_usecase_dsp_base extends seq_dsp_base_kat;
  task body();
    `uvm_do(init_seq)                    // init clock, power and reset
    `uvm_do_with(config_xxx_seq,{...})   // execute all config seqs
    load_adc_data(scenario);             // load raw ADC data from scenario file
    `uvm_do_with(trigger_seq,{...})      // trigger radar operation
    `uvm_do_with(interrupt_seq,{...})    // wait with calculated timeout
    `uvm_do_with(result_seq,{...})       // check results in registers
    `uvm_do_with(status_seq,{...})       // check status in registers
```
(12)

```
class seq_usecase_xxx_kat extends seq_usecase_dsp_base;  // where xxx = scenario
  constraint xxx_c {m_scenario == SCEN_XXX;} // full DSP chain on selected scenario
```
(13)

## IX. CONCLUSION

In this paper we have presented several techniques for making verification of complex mathematical functions more effective including usage of design-for-verification modes, pragmatic application of known-answer-tests, leveraging selected data scenario sets and delayed attention to mathematical modeling. These techniques have emerged through application in various projects and were discussed in the context of single-chip radar devices using a UVM testbench environment. By applying these techniques, we were able to ensure a linear development of the hardware design by identifying bugs early and regularly throughout the project phases, ensuring effective and pragmatic verification results.